\documentclass[10pt,twocolumn,letterpaper]{article}
\usepackage[OT1]{fontenc}
\usepackage[pagenumbers]{cvpr}

\usepackage{afterpage}
\newcommand{\tacitQueueLaterResults}{%
  \ifnum\value{page}<3\relax
    \afterpage{\tacitQueueLaterResults}%
  \else
    \begin{table*}[t]
\centering
\small
\caption{\textbf{Manipulation success (\%).} (a) UniVTAC simulation; (b) real-world tasks. Train/Test Tac.: tactile access during training/inference. HapticVLA and TacImag use the $\pi_{0.5}$/SimpleTouch backbone. Bold: best without measured touch at inference.}
\label{tab:main}\label{tab:real_robot}
\setlength{\tabcolsep}{2pt}
\newcommand{\tacitMainHeader}[1]{\makebox[\dimexpr(\textwidth-0.135\textwidth-0.153\textwidth-16\tabcolsep)/7\relax]{\textbf{#1}}}
\begin{tabular}{@{}ll*{7}{c}@{}}
\toprule
\multicolumn{9}{@{}l@{}}{\textbf{(a) Simulation (UniVTAC)}} \\
\midrule
\makebox[0.135\textwidth][c]{\textbf{Category}} & \makebox[0.153\textwidth][c]{\textbf{Method}} & \tacitMainHeader{Train Tac.} & \tacitMainHeader{Test Tac.} & \tacitMainHeader{Lift Bottle} & \tacitMainHeader{Lift Can} & \tacitMainHeader{Put Bottle} & \tacitMainHeader{Insert Tube} & \tacitMainHeader{Avg.} \\
\midrule
\raisebox{-5.5pt}[0pt][0pt]{\textbf{Vision-only}} & \textbf{ACT}~\cite{zhao2023act} & $\times$ & $\times$ & 42 & 20 & 28 & 45 & 33.8 \\
 & $\boldsymbol{\pi_{0.5}}$~\cite{pi05} & $\times$ & $\times$ & 92 & 59 & 31 & 64 & 61.5 \\
\midrule
\raisebox{-16.5pt}[0pt][0pt]{\textbf{Vision + Tac.}} & \textbf{VITaL}~\cite{vital} & $\checkmark$ & $\checkmark$ & 72 & 8 & 32 & 34 & 36.5 \\
 & \textbf{UniVTAC-ACT}~\cite{univtac} & $\checkmark$ & $\checkmark$ & 71 & 29 & 31 & 56 & 46.8 \\
 & \textbf{FTP-1}~\cite{ftp1} & $\checkmark$ & $\checkmark$ & 97 & 65 & 47 & 79 & 72.0 \\
 & \textbf{SimpleTouch}~\cite{simpletouch} & $\checkmark$ & $\checkmark$ & 100 & 68 & 50 & 100 & 79.5 \\
\midrule
\raisebox{-11pt}[0pt][0pt]{\textbf{Train-time Tac.}} & \textbf{HapticVLA}~\cite{hapticvla} & $\checkmark$ & $\times$ & 97 & 61 & 35 & 69 & 65.5 \\
 & \textbf{TacImag}~\cite{tacimag} & $\checkmark$ & $\times$ & \textbf{100} & 63 & \textbf{45} & 87 & 73.8 \\
 & \textbf{\mbox{TacitTouch}} & $\checkmark$ & $\times$ & 96 & \textbf{85} & 37 & \textbf{92} & \textbf{77.5} \\
\midrule[0.6pt]
\multicolumn{9}{@{}l@{}}{\textbf{(b) Real world}} \\
\midrule
\makebox[0.135\textwidth][c]{\textbf{Category}} & \makebox[0.153\textwidth][c]{\textbf{Method}} & \tacitMainHeader{Train Tac.} & \tacitMainHeader{Test Tac.} & \tacitMainHeader{Ball Sorting} & \tacitMainHeader{Insert USB} & \tacitMainHeader{Wipe Board} & \tacitMainHeader{Chip Moving} & \tacitMainHeader{Avg.} \\
\midrule
\textbf{Vision-only} & $\boldsymbol{\pi_{0.5}}$~\cite{pi05} & $\times$ & $\times$ & 75 & 5 & 65 & 55 & 50.00 \\
\midrule
\textbf{Vision + Tac.} & \textbf{SimpleTouch}~\cite{simpletouch} & $\checkmark$ & $\checkmark$ & 100 & 25 & 90 & 90 & 76.25 \\
\midrule
\raisebox{-5.5pt}[0pt][0pt]{\textbf{Train-time Tac.}} & \textbf{TacImag}~\cite{tacimag} & $\checkmark$ & $\times$ & 55 & 15 & 75 & 75 & 55.00 \\
 & \textbf{\mbox{TacitTouch}} & $\checkmark$ & $\times$ & \textbf{80} & \textbf{25} & \textbf{85} & \textbf{100} & \textbf{72.50} \\
\bottomrule
\end{tabular}
\end{table*}
    \begin{table*}[t]
\centering
\caption{\textbf{Visual robustness on UniVTAC (\%).} Clean: unperturbed; Light: half camera-image brightness; Camera: random pose shifts including occlusion. Bold: best without measured touch at inference.
Same 100 scheduled seeds per method and task; errors are excluded. Only SimpleTouch receives measured tactile input at inference.
Mean changes from Clean (Light / Camera, percentage points): $\pi_{0.5}$: 0.0 / $-$11.5; SimpleTouch: $-$1.8 / $-$14.5;
TacImag: \textbf{$-$9.3 / $-$30.5}; \mbox{TacitTouch}: \textbf{0.0 / $-$19.0}.}
\label{tab:robustness_delta}
\begingroup
\small
\setlength{\tabcolsep}{1.2pt}
\renewcommand{\arraystretch}{1.13}
\newcommand{\robustTaskHeader}[2]{\parbox[c]{\dimexpr(\linewidth-0.13\linewidth-0.186\linewidth-30\tabcolsep)/12\relax}{\centering\strut\textbf{#1}\\\textbf{#2}\strut}}
\newcommand{\robustAvgHeader}{\parbox[c]{0.062\linewidth}{\centering\strut\textbf{Avg.}\strut}}
\begin{tabular}{@{}l*{15}{c}@{}}
\toprule
 & \multicolumn{5}{c}{\textbf{Clean}} & \multicolumn{5}{c}{\textbf{Light}} & \multicolumn{5}{c}{\textbf{Camera}} \\
\cmidrule(lr){2-6}\cmidrule(lr){7-11}\cmidrule(lr){12-16}
\parbox[c]{0.13\linewidth}{\centering\strut\textbf{Method}\strut} & \robustTaskHeader{Lift}{Bottle} & \robustTaskHeader{Lift}{Can} & \robustTaskHeader{Put}{Bottle} & \robustTaskHeader{Insert}{Tube} & \robustAvgHeader & \robustTaskHeader{Lift}{Bottle} & \robustTaskHeader{Lift}{Can} & \robustTaskHeader{Put}{Bottle} & \robustTaskHeader{Insert}{Tube} & \robustAvgHeader & \robustTaskHeader{Lift}{Bottle} & \robustTaskHeader{Lift}{Can} & \robustTaskHeader{Put}{Bottle} & \robustTaskHeader{Insert}{Tube} & \robustAvgHeader \\
\midrule
$\boldsymbol{\pi_{0.5}}$ & 92 & 59 & 31 & 64 & 61.5 & 89 & 59 & 28 & 70 & 61.5 & 85 & 41 & 16 & \textbf{58} & 50.0 \\
\midrule
\textbf{SimpleTouch} & 100 & 68 & 50 & 100 & 79.5 & 100 & 65 & 47 & 99 & 77.8 & 100 & 38 & 23 & 99 & 65.0 \\
\midrule
\textbf{TacImag} & \textbf{100} & 63 & \textbf{45} & 87 & 73.8 & 72 & 64 & 24 & \textbf{98} & 64.5 & 85 & 37 & \textbf{30} & 21 & 43.3 \\
\textbf{\mbox{TacitTouch}} & 96 & \textbf{85} & 37 & \textbf{92} & \textbf{77.5} & \textbf{95} & \textbf{86} & \textbf{41} & 88 & \textbf{77.5} & \textbf{92} & \textbf{64} & 20 & \textbf{58} & \textbf{58.5} \\
\bottomrule
\end{tabular}
\endgroup
\end{table*}
    \begin{figure*}[t]
\centering
\includegraphics[width=\textwidth]{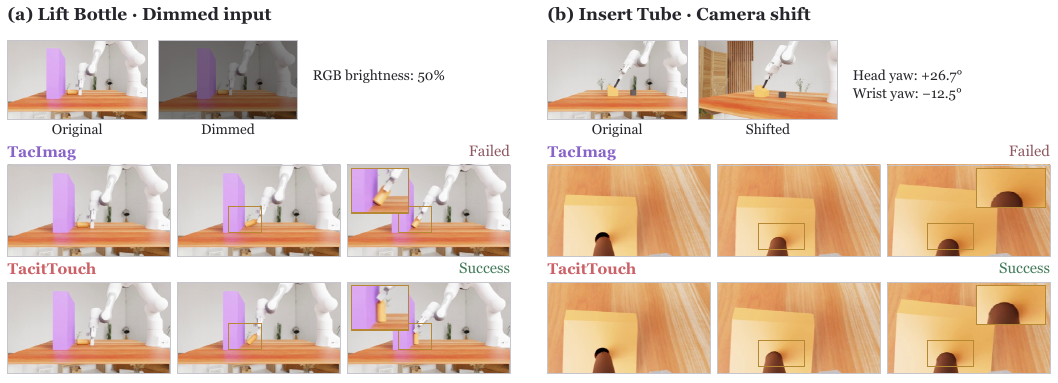}
\caption{\textbf{Contact-rich behavior under visual perturbations.} Selected matched TacImag/\mbox{TacitTouch} rollouts under (a) dimmed input and (b) shifted cameras. Top pairs show original/perturbed head views; trajectories use head/wrist views, respectively. Insets enlarge contact. Dimming affects policy inputs, not trajectory colors; selection details: Appendix~\ref{sec:perturbation_reproducibility}.}
\label{fig:behavior_perturbations}
\end{figure*}
    \begin{figure*}[t]
\centering
\includegraphics[width=\textwidth]{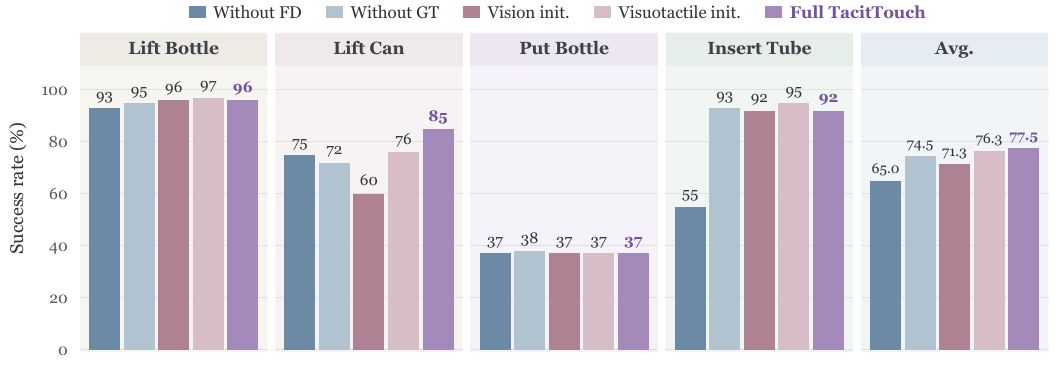}
\caption{\textbf{Components of tactile-guided transfer.} Variants share a frozen tactile teacher and 20,000 updates; task-trained initializations inherit prior training. FD: feature distillation; GT: demonstrated-action supervision. Full values: Appendix~\ref{sec:plot_values}.}
\label{fig:ablations}
\end{figure*}
    \begin{figure*}[t]
\centering
\includegraphics[width=\textwidth]{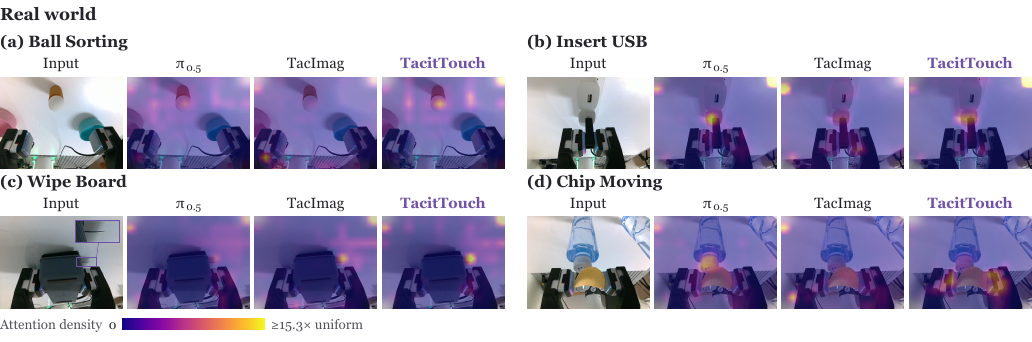}
\caption{\textbf{Task-relevant visual attention on real-world tasks.} Matched wrist observations and attention maps from $\pi_{0.5}$, TacImag, and \mbox{TacitTouch}. Warmer colors indicate higher attention density relative to uniform, on a common scale. The inset in Wipe Board enlarges residual writing in the original input.}
\label{fig:visual_sensitivity}
\end{figure*}
  \fi
}
\newcounter{tacitSavedDblTopNumber}
\definecolor{cvprblue}{rgb}{0.21,0.49,0.74}
\usepackage[pagebackref,breaklinks,colorlinks,allcolors=cvprblue]{hyperref}
\title{Distilling Touch into Vision-Only Policies for Contact-Rich Manipulation}
\author{%
  Chen Yang\textsuperscript{1,}\thanks{These authors contributed equally.}\quad Linhao Wang\textsuperscript{1,}\footnotemark[1]\quad Linzhe Shi\textsuperscript{2}\quad Guohui Zhang\textsuperscript{1}\quad Changjie Wu\textsuperscript{2}\\
  Hang Zhang\textsuperscript{2}\quad Haoyu Li\textsuperscript{1}\quad Mu Xu\textsuperscript{2}\quad Jiansheng Fan\textsuperscript{1}\quad Chen Wang\textsuperscript{1,}\thanks{Corresponding author.}\\[0.5ex]
  \textsuperscript{1}Tsinghua University\qquad \textsuperscript{2}Amap, Alibaba Group%
}
\hypersetup{%
  pdftitle={Distilling Touch into Vision-Only Policies for Contact-Rich Manipulation},
  pdfauthor={Chen Yang, Linhao Wang, Linzhe Shi, Guohui Zhang, Changjie Wu, Hang Zhang, Haoyu Li, Mu Xu, Jiansheng Fan, Chen Wang}%
}
\begin{document}
\maketitle
\begin{abstract}
Tactile feedback benefits contact-rich manipulation, yet deploying tactile sensors introduces additional hardware and maintenance requirements.
Tactile experience can still provide effective supervision for policies that operate without tactile sensors at deployment.
Tactile imagination methods such as TacImag remove this hardware dependency by predicting tactile observations, but retain a reconstruction stage whose errors can propagate into action generation.
We present \mbox{TacitTouch}, a distillation framework that transfers tactile-guided behavior directly into a vision-only policy.
The core shift is from ``learn to imagine what touch would feel like'' to ``learn to behave as if you could feel touch'': the student directly learns tactile-guided action patterns without reconstructing tactile observations.
A frozen visuotactile teacher supervises the student's denoising predictions and intermediate action representations at the same noisy action and flow time.
Training also uses demonstrated-action supervision.
Deployment retains only the visual policy, with no tactile encoder, expert, or generator.
Across four simulated UniVTAC tasks and four real-world tasks, \mbox{TacitTouch} achieves absolute improvements of 16.0\% and 22.5\% in mean success rate over vision-only control, respectively.
The absolute gaps to the teacher using measured tactile input are only 2.0\% and 3.75\%, respectively.
In robustness tests, it also outperforms TacImag, with absolute gains of up to 20.0\% in mean success rate.
Project page: {\normalfont\url{https://pier-lab.github.io/tacittouch/}}
\end{abstract}

\begin{figure}[t]
\centering
\includegraphics[width=\columnwidth]{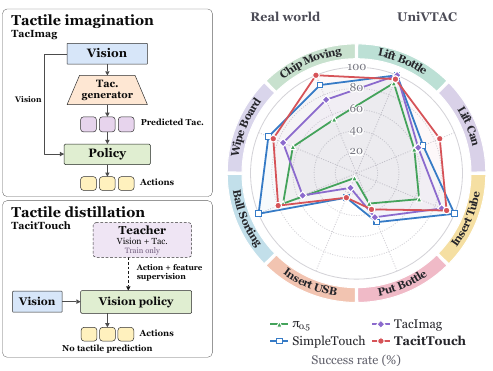}
\caption{\textbf{Imagination versus distillation.} Left: TacImag~\cite{tacimag} predicts touch at inference; \mbox{TacitTouch} distills actions and features during training. Language and state are omitted. Right: success (\%) across four UniVTAC and four real-world tasks.}
\label{fig:comparison}
\end{figure}

\begin{figure*}[t]
\centering
\includegraphics[width=\textwidth]{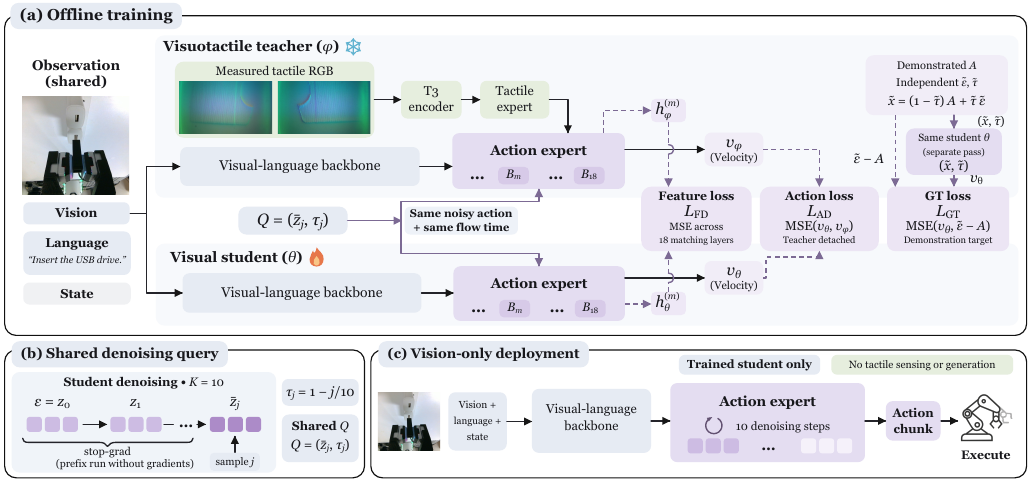}
\caption{\textbf{\mbox{TacitTouch} training and deployment.} (a) A frozen visuotactile teacher supervises student velocities and 18-layer action features at shared noisy actions and flow times; demonstrated actions use a separate pass with independent noise and time. (b) Queries follow the student's ten-step denoising trajectory. (c) Deployment retains only the visual backbone and action expert.}
\label{fig:overview}
\end{figure*}

\begin{figure*}[t]
\centering
\includegraphics[width=\textwidth]{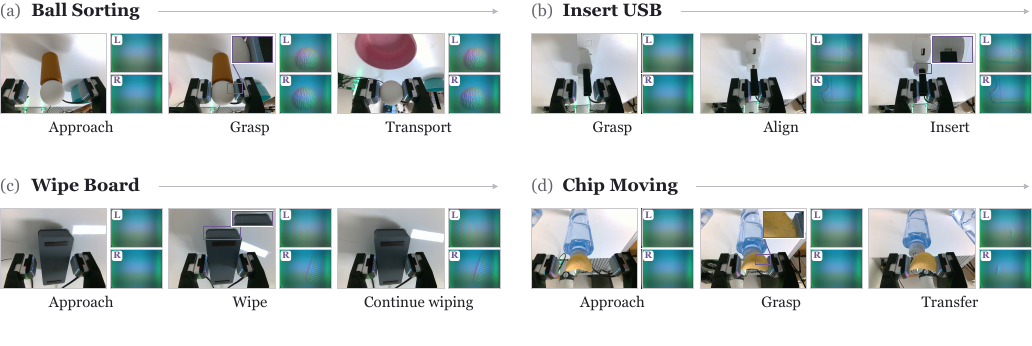}
\caption{\textbf{Real-world visuotactile task sequences.} Demonstration wrist images and paired left/right TacRGB for (a) Ball Sorting, (b) Insert USB, (c) Wipe Board, and (d) Chip Moving, from left to right. Purple insets enlarge interaction regions.}
\label{fig:real_robot_tasks}
\end{figure*}

\tacitQueueLaterResults

\section{Introduction}
\label{sec:introduction}
Contact-rich manipulation depends on physical interactions that are only partly visible.
A grasp can appear secure while an object slips; an insertion can appear aligned while contact blocks further motion.
Pretrained vision-language-action policies provide strong visual and language priors~\cite{pi05}, while tactile sensing supplies complementary contact information~\cite{ftp1,simpletouch}.
Tactile feedback benefits contact-rich manipulation, yet deploying tactile sensors introduces additional hardware and maintenance requirements.
We therefore ask whether tactile experience can strengthen a vision-only policy.

Tactile imagination offers one route to this goal: TacImag~\cite{tacimag} predicts tactile observations from vision and state to guide action generation.
This removes the need for physical tactile sensors but retains tactile generation and processing along the deployed action path, where prediction errors may affect subsequent actions.
HapticVLA's released procedure trains a visual student using demonstrated and teacher-generated actions~\cite{hapticvla,hapticvlacode}, but does not explicitly distill intermediate action representations.
These representations capture how the teacher uses touch to generate actions and provide a key source of supervision for transferring tactile-guided behavior.

We introduce \textbf{\mbox{TacitTouch}}, a framework that transfers tactile-guided behavior from a frozen SimpleTouch teacher~\cite{simpletouch} to a vision-only student.
The teacher receives measured touch alongside the visual, language, and proprioceptive inputs available to the student.
We jointly match its denoising velocities and intermediate action-expert features to supervise both outputs and their computation.
Following on-policy distillation~\cite{stepopd}, both share a noisy action and flow time sampled from the student's denoising trajectory.
Training uses offline demonstration observations without environment rollouts, with independent flow-matching supervision from demonstrated actions.

This use of interaction experience to guide action generation parallels how humans draw on prior manipulation experience: recognizing familiar objects allows people to scale lifting forces before weight feedback becomes available~\cite{gordon1993memory,chouinard2005anticipation}.
Fingertip forces can also adapt to an object's actual weight while subjective weight judgments remain biased~\cite{flanagan2000predictions}.
The distinction between motor adaptation and perceptual experience offers an intuitive motivation for transferring experience through behavior.

Figure~\ref{fig:comparison} captures this behavioral focus: learn to imagine what touch would feel like versus learn to behave as if you could feel touch.
\mbox{TacitTouch} places tactile guidance in the training objective and retains the student's standard visual control path at deployment, without a tactile sensor, encoder, expert, or generator.
Compared with tactile imagination, this removes both the cost of tactile reconstruction and the dependence of action generation on its predicted output.

On four contact-rich UniVTAC tasks~\cite{univtac}, \mbox{TacitTouch} reaches 77.5\% mean success, compared with 61.5\% for vision-only control and 79.5\% for the live-touch teacher.
On four real-world tasks, mean success increases from 50.00\% to 72.50\% over vision-only control.
Deployment timing shows that \mbox{TacitTouch} reduces policy-call latency by 25.8\% relative to TacImag.
Its mean success is also higher than that baseline under image-brightness and camera-pose perturbations.

Our contribution is a tactile-to-visual distillation framework that transfers a visuotactile teacher's behavior and action representations into a vision-only policy through joint denoising velocity and action-feature supervision.
Tactile experience thereby directly shapes the student's behavior learning.
The framework retains the student's standard visual control path without tactile sensing, reconstruction, or additional tactile processing at deployment.
Simulation and real-world experiments demonstrate its benefits for contact-rich manipulation and higher success under visual perturbations.

\section{Related Work}
\label{sec:related}
\paragraph{Visuotactile Policy Learning.}
Visual policies span action-chunk prediction~\cite{zhao2023act,diffusionpolicy}, scalable Transformer control~\cite{rt1,rt2}, and generalist robot models~\cite{octo,openvla,pi0,pi05}.
Optical sensors such as GelSight and DIGIT~\cite{gelsight,digit} expose local contact information.
GelSight-based policies learn grasp and regrasp adjustments~\cite{morefeeling}, while combined visual and haptic representations support insertion~\cite{makingsense}.
Human-collected paired data~\cite{touchgo} and multimodal, multisensor learning~\cite{unitouch,t3,anytouch2} support tactile representations across settings.
FTP-1 develops cross-sensor tactile policies~\cite{ftp1}, while VITaL pretrains policies with and without deployment-time touch~\cite{vital}.
SimpleTouch combines a frozen tactile encoder with a task-trained tactile expert~\cite{simpletouch}; Dream-Tac jointly models actions, future vision, and tactile dynamics conditioned on current touch~\cite{lou2026dreamtac}.
UniVTAC supplies a common simulation benchmark~\cite{univtac}.
We use task-trained SimpleTouch as a teacher to transfer tactile-guided behavior while retaining the student's visual architecture.

\paragraph{Tactile Knowledge Transfer.}
Learning with privileged information uses training-only signals to guide models with restricted inputs~\cite{vapnik2009,vapnik2015,generalizeddistillation}.
Robotic approaches exploit privileged state through asymmetric critics~\cite{asymactor}, teacher-trained visual driving policies~\cite{cheating}, and learned motor-adaptation modules~\cite{rma}.
For tactile transfer, TacImag predicts touch from vision and state, retaining tactile generation and processing at deployment~\cite{tacimag}.
HapticVLA's released action-mixing procedure trains visual students with demonstrated and teacher-generated actions~\cite{hapticvlacode,hapticvla}.
\mbox{TacitTouch} jointly supervises denoising predictions and intermediate action representations, transferring tactile-guided behavior without tactile reconstruction at deployment.

\paragraph{Knowledge Distillation.}
Output distillation~\cite{hinton2015} extends to control~\cite{rusu2016}, while intermediate-feature supervision~\cite{fitnets} supports cross-modal transfer~\cite{gupta2016}.
Other representation targets include attention~\cite{attentiontransfer}, inter-sample relations~\cite{rkd}, contrastive representations~\cite{crd}, and cross-layer feature review~\cite{reviewkd}.
Generative models accelerate sampling via progressive~\cite{progressivedistillation}, consistency~\cite{consistencymodels}, and distribution-matching distillation~\cite{dmd,dmd2}.
STEP-OPD aligns inter-block representation changes and uses extrapolated velocities in student-trajectory distillation~\cite{stepopd}.
We use ordinary on-policy velocity matching at shared noisy actions and flow times, directly aligning corresponding action-expert hidden states with mean squared error.

\section{Method}
\label{sec:method}
\mbox{TacitTouch} distills a visuotactile teacher's action computation into a visual student, allowing tactile experience to guide policy learning while preserving the original visual policy's deployment architecture.
Figure~\ref{fig:overview} shows the training and deployment pipeline.

\subsection{Problem Formulation}
Given a visual policy $\pi_V$, we train an architecturally identical student $\pi_\theta$ using action supervision from a visuotactile teacher $\pi_\phi$ with access to measured touch.
The student retains the original backbone, action generation module, and input and output interfaces.

A demonstration dataset $\mathcal{D}$ contains synchronized camera observations $I_t$, a language instruction $\ell$, proprioceptive state $s_t$, measured tactile observations $U_t$, and demonstrated action chunks $A_t=(a_t,\ldots,a_{t+H-1})$ of length $H$.
The teacher and student observations are
\begin{equation}
o_t^{VT}=(I_t,\ell,s_t,U_t),
\qquad
o_t^V=(I_t,\ell,s_t).
\label{eq:observations}
\end{equation}
The teacher generates actions from $o_t^{VT}$ and the student from $o_t^V$, with
\begin{equation}
\operatorname{Arch}(\pi_\theta)=\operatorname{Arch}(\pi_V).
\label{eq:architecture}
\end{equation}
Here, \emph{vision-only} denotes the absence of tactile input; language and proprioception remain available to the base policy.

This formulation is not tied to a particular visual backbone or tactile encoding scheme.
Distillation requires comparable action predictions and intermediate representations between the teacher and student.

\subsection{Tactile-Guided Distillation}
\label{sec:action_distillation}
\mbox{TacitTouch} uses on-policy distillation~\cite{stepopd} to supervise action outputs and intermediate representations along the student's denoising trajectory, alongside an independent demonstrated-action objective.

\paragraph{Shared Denoising Query.}
Both policies predict action velocities through conditional flow matching~\cite{lipman2023}, denoted by $v_\phi$ and $v_\theta$, with flow time $\tau=1$ corresponding to noise and $\tau=0$ to clean actions.
For an observation drawn from $\mathcal{D}$, sample Gaussian noise $\epsilon\sim\mathcal{N}(0,\mathbf{I})$ and a query index $j\sim\operatorname{Uniform}\{0,\ldots,K-1\}$.
Starting from $z_0=\epsilon$, set $\tau_k=1-k/K$ and execute the first $j$ student Euler steps without gradients:
\begin{equation}
z_{k+1}=z_k-\frac{1}{K}v_\theta(z_k,\tau_k\mid o_t^V),
\quad k=0,\ldots,j-1.
\label{eq:euler}
\end{equation}
The detached state $\bar{z}_j=\operatorname{stopgrad}(z_j)$ defines the shared query $Q_j=(\bar{z}_j,\tau_j)$.
Teacher and student receive the same noisy action, flow time, and non-tactile observations; only the teacher additionally receives measured touch.

\paragraph{Action Distillation.}
At the shared query, we match teacher and student denoising velocities:
\begin{equation}
\begin{aligned}
\mathcal{L}_{\mathrm{AD}}=
\mathbb{E}\bigl[&\bigl\|
v_\theta(\bar{z}_j,\tau_j\mid o_t^V)\\
&-\operatorname{stopgrad}\!\left(v_\phi(\bar{z}_j,\tau_j\mid o_t^{VT})\right)
\bigr\|_{\mathrm{MSE}}^2\bigr].
\end{aligned}
\label{eq:action_distillation}
\end{equation}
Here, $\|\cdot\|_{\mathrm{MSE}}^2$ averages squared differences over all elements of the action tensor.
Neither the teacher targets nor the student prefix that generates the query participates in backpropagation; gradients update the student only through its predictions at the shared query.

\paragraph{Action-Feature Distillation.}
Velocity matching constrains action outputs but does not explicitly align the intermediate representations that generate them.
We therefore also match teacher and student action features at the same query.
Let $h_\theta^{(m)}$ and $h_\phi^{(m)}$ denote the action-token hidden states output by the $m$th corresponding action layer.
We minimize
\begin{equation}
\mathcal{L}_{\mathrm{FD}}=
\mathbb{E}\left[
\frac{1}{M}\sum_{m=1}^{M}
\left\|
h_\theta^{(m)}-
\operatorname{stopgrad}\!\left(h_\phi^{(m)}\right)
\right\|_{\mathrm{MSE}}^2
\right].
\label{eq:feature_distillation}
\end{equation}
We extract the hidden states after each Transformer block and before the final action-expert normalization, averaging over samples, layers, action tokens, and hidden channels.
Corresponding features have the same shape, allowing direct comparison of raw hidden states without a learned projection, feature normalization, or differences between layers.
This objective supervises the visual student with the teacher's tactile-conditioned action representations.

\paragraph{Demonstrated-Action Supervision.}
The base visual policy's flow-matching objective provides direct supervision from demonstrated actions.
A separate student forward pass draws independent Gaussian noise $\tilde{\epsilon}$ and flow time $\tilde{\tau}$, both independent of the distillation query.
Flow time follows the base policy's supervised sampling distribution rather than the discrete student-trajectory distribution.
We construct
\begin{equation}
\tilde{x}_{\tilde{\tau}}=(1-\tilde{\tau})A_t+\tilde{\tau}\tilde{\epsilon},
\end{equation}
and optimize
\begin{equation}
\mathcal{L}_{\mathrm{GT}}=
\mathbb{E}\left[
\left\|
v_\theta(\tilde{x}_{\tilde{\tau}},\tilde{\tau}\mid o_t^V)
-(\tilde{\epsilon}-A_t)
\right\|_{\mathrm{MSE}}^2
\right].
\label{eq:ground_truth}
\end{equation}

The full training objective is
\begin{equation}
\mathcal{L}=\mathcal{L}_{\mathrm{AD}}
+\lambda_{\mathrm{FD}}\mathcal{L}_{\mathrm{FD}}
+\lambda_{\mathrm{GT}}\mathcal{L}_{\mathrm{GT}}.
\label{eq:full_objective}
\end{equation}
Our flow-matching implementation uses all $M=18$ action-expert layers, with default weights $\lambda_{\mathrm{FD}}=0.01$ and $\lambda_{\mathrm{GT}}=1$.

\subsection{Training and Deployment}
\label{sec:training_deployment}
We use frozen, task-trained SimpleTouch~\cite{simpletouch} as the teacher, initialize the student from pretrained $\pi_{0.5}$ weights~\cite{pi05}, and jointly train its visual-language backbone and action expert.
Offline training uses synchronized demonstrations without environment rollouts; teacher and student share augmented images, action representations, and normalization.

Main simulation training uses eight-GPU FSDP~\cite{fsdp}, global batch size 64, and 20,000 AdamW~\cite{adamw} updates.
We evaluate the final checkpoint; learning-rate schedules and further settings appear in Appendix~\ref{sec:training_details}.

At deployment, the student uses current images, instructions, and proprioception to generate a 50-action chunk from Gaussian noise through ten Euler steps (Eq.~\ref{eq:euler}).
The base policy's inverse normalization and action adapter produce control commands; task settings determine how many actions execute before the next observation.
No teacher or tactile processing is used.

\section{Experiments}
\label{sec:experiments}
We evaluate manipulation performance, visual robustness, and deployment cost, then use ablations and analysis to examine the effects and scope of tactile-guided transfer.

\subsection{Experimental Setup}
\label{sec:setup}
\paragraph{Tasks and Platforms.}
The main simulation experiments use four UniVTAC tasks~\cite{univtac}: Lift Bottle, Lift Can, Put Bottle, and Insert Tube.
Real-world experiments use a Franka FR3 robot, GelSight Mini tactile sensors, and a wrist-mounted D405 camera to evaluate Ball Sorting, Insert USB, Wipe Board, and Chip Moving; representative task sequences appear in Fig.~\ref{fig:real_robot_tasks}.
Our trained configurations use the same 50 demonstrations per task.
Four additional simulation tasks examine transfer scope in Sec.~\ref{sec:transfer_scope}.

\paragraph{Baselines.}
The primary comparisons share the $\pi_{0.5}$/SimpleTouch backbone~\cite{pi05,simpletouch}, including vision-only control, measured-touch SimpleTouch, TacImag~\cite{tacimag}, and \mbox{TacitTouch}.
Simulation additionally includes HapticVLA~\cite{hapticvla,hapticvlacode}.
Among these methods, SimpleTouch receives measured touch at deployment, TacImag uses predicted touch, and HapticVLA and \mbox{TacitTouch} deploy visual students directly.
Additional ACT, VITaL, UniVTAC-ACT, and FTP-1 results come from the SimpleTouch benchmark study~\cite{simpletouch}; implementation details are provided in the appendix.

\paragraph{Evaluation Protocol.}
Simulation schedules 100 episodes per method and task, excluding runtime and video-decoding errors from the valid denominator; real-world evaluation uses 20 rollouts per method and task.
We report task success rates and their unweighted mean, evaluating the final training checkpoints.
Section~\ref{sec:training_deployment} describes training; environment versions, camera configurations, random seeds, and full evaluation details are given in the appendix.

\subsection{Manipulation Performance}
\label{sec:main_results}
\label{sec:real_robot}
\paragraph{Simulation Results.}
\mbox{TacitTouch} improves all four UniVTAC tasks over vision-only control (Table~\ref{tab:main}(a)), raising mean success from 61.5\% to 77.5\%; Lift Can and Insert Tube gain most.
In mean success, it leads shared-backbone methods deployed without measured touch and approaches SimpleTouch.
TacImag outperforms the student on Lift Bottle and Put Bottle; the student exceeds the teacher on Lift Can.

\paragraph{Real-World Results.}
\mbox{TacitTouch} also outperforms vision-only control and TacImag on every physical task, raising the vision baseline's mean from 50.00\% to 72.50\% (Table~\ref{tab:main}(b)).
Chip Moving gives the largest improvement; teacher and student tie on Insert USB.

\begin{table}[t]
\centering
\small
\caption{\textbf{Real-world brightness robustness (\%).} Camera RGB $\times0.5$; 20 rollouts per method/task. Bold: best without measured touch at inference.}
\label{tab:real_robot_light}
\begingroup
\setlength{\tabcolsep}{1.5pt}
\renewcommand{\arraystretch}{1.13}
\newcommand{\tacitRealLightHeader}[1]{\parbox[c]{\dimexpr(\columnwidth-0.27\columnwidth-10\tabcolsep)/5\relax}{\centering\strut\textbf{#1}\strut}}
\begin{tabular}{@{}l*{5}{c}@{}}
\toprule
\parbox[c]{0.27\columnwidth}{\centering\strut\textbf{Method}\strut} & \tacitRealLightHeader{Ball\\Sorting} & \tacitRealLightHeader{Insert\\USB} & \tacitRealLightHeader{Wipe\\Board} & \tacitRealLightHeader{Chip\\Moving} & \tacitRealLightHeader{Avg.} \\
\midrule
$\boldsymbol{\pi_{0.5}}$ & 65 & 5 & 50 & 35 & 38.75 \\
\midrule
\textbf{SimpleTouch} & 100 & 20 & 85 & 90 & 73.75 \\
\midrule
\textbf{TacImag} & 45 & 5 & 70 & 65 & 46.25 \\
\textbf{\mbox{TacitTouch}} & \textbf{85} & \textbf{25} & \textbf{75} & \textbf{80} & \textbf{66.25} \\
\bottomrule
\end{tabular}
\endgroup
\end{table}

\subsection{Visual Robustness}
\label{sec:robustness}

Visual perturbations are also studied in dedicated VLA benchmarks~\cite{liberoplus}.
We evaluate fixed checkpoints against unperturbed Clean.
Light halves camera RGB values before model preprocessing in simulation and the real world, leaving scene lighting and tactile preprocessing unchanged.
Simulation-only Camera uniformly shifts head-camera yaw within $\pm30^\circ$ and, when used, wrist-camera yaw within $\pm15^\circ$.
Methods share environment seeds and per-episode perturbations, including views that cause occlusion.

\mbox{TacitTouch} outperforms TacImag under brightness perturbations on both platforms.
Its simulation mean remains 77.5\%, compared with 64.5\% for TacImag (Table~\ref{tab:robustness_delta}).
Its real-world mean is 66.25\%, compared with 46.25\% for TacImag (Table~\ref{tab:real_robot_light}).
On the real robot, \mbox{TacitTouch} exceeds both vision-only control and TacImag on all four tasks and shows a smaller decrease from Clean.
SimpleTouch retains the highest mean on both platforms, while TacImag remains stronger on simulated Insert Tube.
Figure~\ref{fig:behavior_perturbations}(a) shows a Lift Bottle case with matched seeds and perturbations: \mbox{TacitTouch} completes the lift, while TacImag fails.
Trajectory frames retain their original recorded brightness because dimming is applied in the policy-input path.

In TacImag's two-stage inference path, tactile-prediction errors can propagate into subsequent action generation; \mbox{TacitTouch} generates actions directly without this intermediate prediction, providing a plausible explanation for the robustness gains.

Camera perturbations reduce mean success for every method.
\mbox{TacitTouch} reaches 58.5\% versus 43.3\% for TacImag (Table~\ref{tab:robustness_delta}).
The largest task-level margin appears on Insert Tube: \mbox{TacitTouch} succeeds in 58\% of episodes, versus 21\% for TacImag.
This gap exceeds the difference in Clean, where \mbox{TacitTouch} and TacImag achieve 92\% and 87\%, respectively.
SimpleTouch remains stronger overall, while TacImag leads on Put Bottle.
Figure~\ref{fig:behavior_perturbations}(b) shows an Insert Tube comparison under matched conditions: \mbox{TacitTouch} completes insertion, while TacImag fails.
These selected cases illustrate behavioral differences.

\begin{table}[t]
\centering
\small
\caption{\textbf{Inference latency (s/call).} Mean policy-call time; A100 GPU, batch one, ten flow steps.}
\label{tab:latency}
\setlength{\tabcolsep}{1.5pt}
\newcommand{\tacitLatencyHeader}[1]{\parbox[c]{\dimexpr(\columnwidth-0.27\columnwidth-10\tabcolsep)/5\relax}{\centering\strut\textbf{#1}\strut}}
\begin{tabular}{@{}l*{5}{c}@{}}
\toprule
\parbox[c]{0.27\columnwidth}{\centering\strut\textbf{Method}\strut} & \tacitLatencyHeader{Lift\\Bottle} & \tacitLatencyHeader{Lift\\Can} & \tacitLatencyHeader{Put\\Bottle} & \tacitLatencyHeader{Insert\\Tube} & \tacitLatencyHeader{Avg.} \\
\midrule
\textbf{Vision + Tac.} & 0.372 & 0.379 & 0.360 & 0.376 & 0.372 \\
\textbf{TacImag} & 0.452 & 0.456 & 0.452 & 0.455 & 0.454 \\
\textbf{\mbox{TacitTouch}} & \textbf{0.336} & \textbf{0.337} & \textbf{0.334} & \textbf{0.340} & \textbf{0.337} \\
\bottomrule
\end{tabular}
\end{table}

\subsection{Deployment Efficiency}
\label{sec:latency}

After stable warmup, we measure 100 calls per method and task, synchronizing the GPU before and after each call.
The timed interval includes preprocessing, model computation, and CPU action output, excluding checkpoint loading, input decoding, simulation, and communication.
The complete protocol is given in Appendix~\ref{sec:timing}.

\mbox{TacitTouch} achieves 0.337 seconds per call, reducing mean latency by 25.8\% relative to TacImag and 9.5\% relative to SimpleTouch (Table~\ref{tab:latency}).
The deployed model contains 3.35B parameters and retains only the visual-language backbone and action expert.
Timing uses GT0 student checkpoints ($\lambda_{\mathrm{GT}}=0$, without demonstrated-action supervision), while the main performance results use GT1 checkpoints ($\lambda_{\mathrm{GT}}=1$, with this supervision).
Both share the same inference architecture, so the measurements characterize their common deployment cost.

\begin{table}[t]
\centering
\small
\setlength{\tabcolsep}{2pt}
\newcommand{\tacitModalityHeader}[2]{\multicolumn{1}{#1}{{\bfseries\begin{tabular}[c]{@{}c@{}}#2\end{tabular}}}}
\caption{\textbf{Teacher and student modalities (\%).} V: vision, language, and state; VT: V plus measured tactile input. A dash denotes a directly trained reference. The middle rows use matched distillation-only objectives to isolate teacher modality through V$\rightarrow$V versus VT$\rightarrow$V; the final row uses the full objective.}
\label{tab:modality}
\begin{tabular*}{\columnwidth}{@{\extracolsep{\fill}}ccc*{5}{r}@{}}
\toprule
\tacitModalityHeader{@{}c}{Teacher\\input} & \tacitModalityHeader{c}{Student\\input} & \tacitModalityHeader{c}{GT\\loss} & \tacitModalityHeader{c}{Lift\\Bottle} & \tacitModalityHeader{c}{Lift\\Can} & \tacitModalityHeader{c}{Put\\Bottle} & \tacitModalityHeader{c}{Insert\\Tube} & \tacitModalityHeader{c@{}}{Avg.} \\
\midrule
V & -- & -- & 92 & 59 & 31 & 64 & 61.5 \\
VT & -- & -- & 100 & 68 & 50 & 100 & 79.5 \\
\midrule
V & V & $\times$ & 95 & 63 & 38 & 68 & 66.0 \\
VT & VT & $\times$ & 100 & 70 & 47 & 99 & 79.0 \\
VT & V & $\times$ & 95 & 72 & 38 & 93 & 74.5 \\
\midrule
VT & V & $\checkmark$ & 96 & 85 & 37 & 92 & 77.5 \\
\bottomrule
\end{tabular*}
\end{table}

\subsection{Ablation Studies}
\label{sec:ablations}
\paragraph{Feature Distillation.}
With action distillation and GT supervision retained, removing feature distillation reduces mean success from 77.5\% to 65.0\%, the largest decrease among the tested variants (Fig.~\ref{fig:ablations}).
The loss ablations fix the teacher, initialization, and update budget.
The largest task-level decrease occurs on Insert Tube, where success falls from 92\% to 55\%.
The decrease supports feature supervision alongside velocity matching.

\paragraph{Demonstrated-Action Supervision.}
Removing the independent demonstration loss gives 74.5\% mean success, with Lift Can falling from 85\% to 72\%.
Success rates on the other three tasks differ by at most 1\% in absolute terms, indicating task-dependent benefits from direct demonstration supervision.

\paragraph{Student Initialization.}
Pretrained base weights yield higher mean success than task-trained vision or shared teacher weights (Fig.~\ref{fig:ablations}).
Both task-trained variants retain the full objective and restart the optimizer for 20,000 distillation updates, while inheriting prior task training.
Teacher initialization omits tactile modules, keeping the student vision-only.

\paragraph{Teacher and Student Modalities.}
Vision-to-vision distillation raises mean success from 61.5\% to 66.0\% (Table~\ref{tab:modality}), showing that distillation itself can provide some benefit even without tactile guidance.
With the visual student architecture, distillation objectives, and update budget held fixed, a visuotactile teacher further raises mean success to 74.5\%.
This absolute gain of 8.5\% supports the additional value of transferring tactile-guided behavior beyond the common distillation procedure.

\subsection{Analysis}
\label{sec:analysis}
\begin{table}[t]
\centering
\small
\caption{\textbf{Tactile benefit and transfer scope on additional tasks (\%).} Bold values highlight the measured-touch policy's advantage on the first three tasks.}
\label{tab:scope}
\begingroup
\setlength{\tabcolsep}{1.3pt}
\newcommand{\tacitScopeHeader}[2]{\parbox[c]{#1}{\centering\strut\textbf{#2}\strut}}
\newcommand{\tacitScopeTaskHeader}[2]{\tacitScopeHeader{\dimexpr(\columnwidth-0.30\columnwidth-0.08\columnwidth-10\tabcolsep)/4\relax}{#1\\#2}}
\begin{tabular}{@{}lrrrrr@{}}
\toprule
\tacitScopeHeader{0.30\columnwidth}{Method} & \tacitScopeTaskHeader{Pull-out}{Key} & \tacitScopeTaskHeader{Insert}{Hole} & \tacitScopeTaskHeader{Grasp}{Classify} & \tacitScopeTaskHeader{Insert}{HDMI} & \tacitScopeHeader{0.08\columnwidth}{Avg.} \\
\midrule
\textbf{Vision only} & 36 & 25 & 48 & 3 & 28.0 \\
\textbf{Vision + Tac.} & \textbf{62} & \textbf{85} & \textbf{99} & 3 & 62.3 \\
\textbf{\mbox{TacitTouch}} & 42 & 25 & 51 & 4 & 30.5 \\
\bottomrule
\end{tabular}
\endgroup
\end{table}

\subsubsection{Task-Relevant Visual Attention}
\label{sec:visual_sensitivity}
Figure~\ref{fig:visual_sensitivity} compares $\pi_{0.5}$, TacImag, and \mbox{TacitTouch} with matched wrist images, instructions, and initial action noise; each retains its native state input and receives no measured touch.

\mbox{TacitTouch} emphasizes ball-surface texture for discrimination in Ball Sorting and the receiving socket for alignment in Insert USB.
In Wipe Board, attention highlights residual writing beside the eraser, enlarged in the input inset, indicating where further cleaning is needed.
In Chip Moving, it emphasizes the gripper fingers adjacent to the chip, highlighting contact geometry relevant to grasp stability.
Because the visual input is matched across methods, the comparison concerns how each policy weights the same scene rather than differences in camera framing.
The highlighted regions span object appearance, the interaction target, and the gripper-object interface.
These patterns are consistent with the transfer of tactile-guided behavior into task-relevant visual weighting, and the attention maps provide qualitative evidence of where the policies attend.

\subsubsection{Scope of Tactile Transfer}
\label{sec:transfer_scope}
The benefit of measured touch does not transfer uniformly across tasks (Table~\ref{tab:scope}).
On Pull-out Key, Insert Hole, and Grasp Classify, tactile input substantially improves the teacher, while the visual student remains close to the vision-only baseline.
For example, measured touch raises Insert Hole success from 25\% to 85\%, but the student remains at 25\%.
These tasks separate effective tactile control from successful vision-only transfer: a capable teacher does not ensure that a visual student recovers its advantage under the evaluated training setup.

On Insert HDMI, all three methods remain at 3--4\% success, indicating that adding measured touch alone does not resolve this task under the evaluated setup.
Here, there is little measured-touch benefit for the student to recover.
These results do not distinguish insufficient observation information from training-objective or optimization effects.

\section{Conclusion}
We present \mbox{TacitTouch}, which transfers tactile-guided behavior into a vision-only policy through joint distillation of a frozen visuotactile teacher's denoising predictions and intermediate action representations, combined with demonstrated-action supervision.
This allows the student to retain its original visual control path without tactile sensing or reconstruction at deployment.
Across four UniVTAC simulation tasks and four real-world tasks, the student achieves absolute gains of 16.0\% and 22.5\% in mean success over the vision-only baseline, respectively, approaching the measured-touch teacher on this metric.
Further comparisons show that, relative to TacImag, \mbox{TacitTouch} achieves higher mean success under visual perturbations and its deployment architecture reduces policy-call latency by 25.8\%.

\section*{LLM Usage Statement}
Under author direction, OpenAI Codex was used to assist with code implementation and debugging, experiment execution and result analysis, literature retrieval, manuscript drafting and revision, and the preparation of plotting scripts. OpenAI's image generation tool was used to assist in refining schematic illustrations. These tools were used in an assistive role, and the authors take full responsibility for the research methods, experimental results, scientific claims, and final content of the paper.

{
  \small
  \bibliographystyle{ieeenat_fullname}
  \bibliography{tacit_refs}
}
\clearpage
\setcounter{dbltopnumber}{\value{tacitSavedDblTopNumber}}
\appendix
\backrefsetup{disable}
\setcounter{table}{0}
\renewcommand{\thetable}{S\arabic{table}}
\renewcommand{\theHtable}{supplement.\arabic{table}}
\setcounter{figure}{0}
\renewcommand{\thefigure}{S\arabic{figure}}
\renewcommand{\theHfigure}{supplement.\arabic{figure}}

\section{Methods and Experimental Details}
\label{sec:protocol}

Sections~\ref{sec:training_details}--\ref{sec:evaluation_protocol} describe training, baselines, and evaluation.
Section~\ref{sec:perturbation_reproducibility} details the visual perturbations and the selection of examples in Fig.~\ref{fig:behavior_perturbations}.
Section~\ref{sec:timing} describes the inference benchmark.
Sections~\ref{sec:ablation_details}--\ref{sec:plot_values} report ablations, additional tasks, and per-task results.
Section~\ref{sec:attribution} describes the real-world demonstrations and attention maps.

\subsection{Training Configuration and Distillation}
\label{sec:training_details}

\subsubsection{Learning-Rate Schedule, Optimization, and Checkpoints}

We train the four main simulation tasks separately on 50 demonstrations per task.
For each task, we freeze a task-trained SimpleTouch teacher, initialize the student from pretrained $\pi_{0.5}$ weights, and jointly optimize its vision-language backbone and action expert.
We denote the configurations with demonstration-supervision weights $\lambda_{\mathrm{GT}}=1$ and $\lambda_{\mathrm{GT}}=0$ by GT1 and GT0, respectively.
Table~\ref{tab:supp_training} gives the GT1 configuration used in the main simulation experiments.
We disable periodic simulation validation during training and evaluate the checkpoint at update 20,000.

\begin{table}[b]
\centering
\caption{\textbf{TacitTouch training configuration for the main simulation experiments (GT1).}}
\label{tab:supp_training}
\begin{tabular}{@{}p{.39\columnwidth}p{.55\columnwidth}@{}}
\toprule
Setting & Value \\
\midrule
Teacher / student initialization & Task-trained SimpleTouch / pretrained $\pi_{0.5}$ \\
Parallelism / processes & FSDP / 8 \\
Microbatch per process / global batch & 8 / 64, with one gradient accumulation step \\
FSDP mixed precision & bfloat16 parameters for computation, float32 gradient reduction and buffers \\
Optimizer & AdamW \\
Peak / final learning rate & $5\times10^{-5}$ / $5\times10^{-6}$ \\
Learning-rate schedule & 1,000 warm-up updates, then cosine decay \\
AdamW $\beta$ and $\epsilon$ & $(0.9,0.95)$, $10^{-8}$ \\
Weight decay / gradient clipping norm & $10^{-10}$ (effectively no weight decay) / 1 \\
Updates / checkpoint interval & 20,000 / 5,000 \\
Feature distillation / demonstration weights & $0.01$ / $1$ \\
Action horizon / inference denoising steps & 50 / 10 \\
\bottomrule
\end{tabular}
\end{table}

For each real-world task, we freeze the task-trained SimpleTouch teacher and initialize the TacitTouch student from pretrained $\pi_{0.5}$ weights.
Training uses FSDP with eight processes, a microbatch of 8 per process, and a global batch of 64.
Computation uses bfloat16 parameters, and gradient reduction and buffers use float32, as in simulation.
The optimizer, learning-rate schedule, and checkpoint interval follow Table~\ref{tab:supp_training}.
Each task uses the 50 demonstrations described in Sec.~\ref{sec:real_demonstration_processing}.
We disable periodic validation and deploy the checkpoint at update 20,000.

\subsubsection{Proprioception, Action Representation, and Data}
\label{sec:supp_state}

Following Sec.~3.1, $s_t$ denotes the proprioceptive state at the current sample time.
The teacher receives images, the instruction, $s_t$, and measured touch.
The student receives the same images, instruction, and $s_t$.

In simulation, both training and evaluation use the current eight-dimensional joint state as proprioceptive input.
The actions are eight-dimensional absolute joint targets.
The first seven entries are arm joint angles in radians, and the eighth is the displacement of one finger in meters.
The joint-position interface receives the denormalized target and writes the eighth entry to both symmetric finger joints.
Section~\ref{sec:simulation_details} specifies the simulation step.

On the robot, the first seven state dimensions are measured joint angles in radians.
The eighth is the measured total gripper opening in meters.
The raw control \texttt{command} specifies absolute targets for the seven arm joints and the total opening.
Matching source timestamps yields the \texttt{action} labels, as described in Sec.~\ref{sec:real_demonstration_processing}.
Contact can leave the measured opening larger than the target, so the eighth action dimension is taken from the command.

For real-world training, we subtract the corresponding joint angles in $s_t$ from the first seven action dimensions and retain the absolute opening in the eighth.
At inference, we denormalize the model output and add the same state angles back to obtain absolute targets.
We compute state and action normalization statistics separately for each task and reuse the checkpoint statistics at inference.
Real-world normalization uses the first and 99th percentiles $q_{01}$ and $q_{99}$ in each dimension as follows.
\[
\tilde{x}=2\frac{x-q_{01}}{q_{99}-q_{01}+10^{-6}}-1.
\]

Each prediction contains 50 actions, with each eight-dimensional action zero-padded to 32 dimensions.
If fewer than 50 actions remain in a demonstration, we repeat its final action.
The repeated actions contribute to demonstration supervision when that loss is enabled.
In the main simulation experiments, action distillation and demonstration supervision both compute mean squared error over all $50\times32$ elements.

\subsubsection{Image Augmentation and the Distillation Update}

Following Sec.~\ref{sec:action_distillation}, the teacher and student make predictions at the same noisy action and flow time.
Training uses offline demonstrations. The query follows the current student's ten-step Euler denoising trajectory, so query selection is on-policy.
Algorithm~S1 gives one GT1 update.

For the main simulation experiments, we resize camera images while preserving their aspect ratios and pad them to $224\times224$ pixels.
For the head image, we sample a crop spanning $95\%$ of the input height and width, draw an in-plane rotation uniformly from $[-5^\circ,5^\circ]$, and resample to the input size.
The wrist image receives color augmentation only.
Each view receives color jitter with probability 0.5. Brightness, contrast, and hue offsets are sampled uniformly from $[-0.3,0.3]$, $[-0.4,0.4]$, and $[-0.1,0.1]$, respectively.
Hue uses a unit interval for a full turn, and saturation stays fixed.
We preprocess the teacher's measured tactile images separately.

The shared augmentation in Sec.~\ref{sec:training_deployment} applies only to the distillation pass.
The teacher and student use the same augmented camera images for $\mathcal{L}_{\mathrm{AD}}$ and $\mathcal{L}_{\mathrm{FD}}$.
To compute $\mathcal{L}_{\mathrm{GT}}$ in GT1, the student samples a fresh augmentation of the same source images for a separate forward pass.

\paragraph{Algorithm S1. One TacitTouch distillation update (main simulation, GT1).}
\begin{enumerate}
    \item Load the demonstration observations and action chunks. Sample one camera augmentation for both teacher and student during distillation.
    \item Draw $\epsilon\sim\mathcal{N}(0,\mathbf{I})$ and set $z_0=\epsilon$. Draw $j$ uniformly from $\{0,\ldots,9\}$ and share $j$ across training processes.
    \item With gradients disabled, run student denoising steps $k=0,\ldots,j-1$, updating $z_{k+1}=z_k-v_\theta(z_k,1-k/10)/10$. For $j=0$, use the initial noise $z_0$ directly.
    \item Query both teacher and student at $(\bar{z}_j,1-j/10)$, using the detached query state $\bar{z}_j$.
    \item Provide the teacher with the visuotactile observation and the student with its non-tactile inputs. Read both velocity predictions and the features from all 18 action-expert layers.
    \item Use the frozen teacher outputs to compute $\mathcal{L}_{\mathrm{AD}}$ and $\mathcal{L}_{\mathrm{FD}}$.
    \item Draw a fresh augmentation of the source images and independently sample noise and flow time. Run a separate student forward pass to compute $\mathcal{L}_{\mathrm{GT}}$.
    \item Update the student with $\mathcal{L}_{\mathrm{AD}}+0.01\mathcal{L}_{\mathrm{FD}}+\mathcal{L}_{\mathrm{GT}}$.
\end{enumerate}

The distribution for $j$ described before Eq.~\ref{eq:euler} is the marginal distribution for each sample.
In the GT1 configuration, each optimizer update shares a single draw of $j$ from rank zero across the global batch.
This gives all processes the same number of gradient-free prefix steps.
Across updates, $j$ is drawn uniformly from $\{0,\ldots,9\}$.

In the main simulation experiments, we use the outputs of all 18 action-expert Transformer blocks before the final normalization.
Teacher and student hidden states have matching shapes, allowing direct elementwise mean squared error averaged over samples, layers, action tokens, and channels.

For $\mathcal{L}_{\mathrm{GT}}$, we independently draw $b\sim\operatorname{Beta}(1.5,1)$ and use flow time $0.001+0.999b$.
We interpolate the demonstrated actions with fresh noise and train the student to predict the noise minus the demonstrated actions.
The gradients from distillation and demonstration supervision contribute to the same student update.

\subsection{Baseline Implementations and Inputs}
\label{sec:baselines}

\subsubsection{\texorpdfstring{$\pi_{0.5}$}{pi0.5} and SimpleTouch}

In Table~\ref{tab:main}, $\pi_{0.5}$ uses images, language, and proprioception.
SimpleTouch adds a tactile branch to the same backbone. A frozen T3 encoder extracts features from the current tactile observation, and a separate tactile expert processes them.
TacitTouch uses SimpleTouch as the teacher and deploys the visual student alone.
Table~\ref{tab:supp_inputs} lists the tactile inputs and modules used at inference.

\begin{table}[t]
\centering
\caption{\textbf{Tactile inputs and modules used at inference.}}
\label{tab:supp_inputs}
\begin{tabular}{@{}p{.23\columnwidth}p{.24\columnwidth}p{.42\columnwidth}@{}}
\toprule
Method & Tactile input & Tactile modules \\
\midrule
$\pi_{0.5}$ & None & None \\
SimpleTouch & Measured RGB & T3 encoder, tactile expert \\
TacImag & Generated RGB & Tactile generator, T3 encoder, tactile expert \\
HapticVLA & None & None \\
TacitTouch & None & None \\
\bottomrule
\end{tabular}
\end{table}

Real-world $\pi_{0.5}$ and SimpleTouch start from the same pretrained $\pi_{0.5}$ checkpoint, use the real-world optimization settings in Sec.~\ref{sec:training_details}, and use the checkpoint at update 20,000.
SimpleTouch randomly initializes the tactile expert and keeps the T3 encoder frozen during training.

Real-world SimpleTouch conditions on current touch and predicts T3 features from future measured touch as an auxiliary objective.
It samples a target every five steps over a 50-step horizon and pools the spatial features to $6\times6$.
It normalizes the targets by a reference root mean square (RMS) from the frozen T3 encoder.
Let $\gamma$ and $\beta$ denote the per-channel scale and bias in the last LayerNorm. Under the assumption of zero mean and unit variance in each channel before the affine transform, each output channel has second moment $\gamma^2+\beta^2$.
The reference RMS is
\[
    \sqrt{\operatorname{mean}_c\!\left(\gamma_c^2+\beta_c^2\right)},
\]
where $c$ indexes the channels.
The auxiliary loss weight is fixed at 0.01 throughout training, with no warm-up or scheduling.
At the end of a trajectory, we repeat the last target.

\subsubsection{Tactile Generator and Action Policy in TacImag}

TacImag trains in two stages.
In the first stage, the generator takes the current images and normalized proprioception and jointly predicts left and right tactile RGB images at the same time, each $320\times240$ pixels.
The simulation generator uses the task-specific camera views in Sec.~\ref{sec:simulation_details}, with a separate ResNet-18 encoder and group normalization for each view.

We downsample the tactile images by a factor of four with bilinear interpolation before feeding them to a conditional one-dimensional U-Net.
The channel widths are 512, 1,024, and 2,048. Time embeddings have dimension 128, the kernel size is 5, and group normalization uses eight groups.
The DDPM predicts the clean sample with a cosine noise schedule of 100 training diffusion steps and 10 inference denoising steps.
The generator trains for 200 epochs per simulation task, as listed in Table~\ref{tab:supp_generator}.
Training uses eight processes with a microbatch of 8 per process, a global batch of 64, and one gradient accumulation step.
AdamW uses a learning rate of $10^{-4}$, $\beta=(0.95,0.999)$, weight decay $10^{-6}$, and 500 warm-up steps followed by cosine decay.

\begin{table}[t]
\centering
\caption{\textbf{TacImag generator training in simulation.} Each task uses eight AMD Instinct MI308 GPUs. Time is cumulative at the final logged training step.}
\label{tab:supp_generator}
\begin{tabular}{lrrrr}
\toprule
Task & \shortstack{Stage 1\\samples} & Epochs & \shortstack{Final\\step} & \shortstack{Time\\(h)} \\
\midrule
Lift Bottle & 15,334 & 200 & 47,800 & 3.9 \\
Lift Can & 9,172 & 200 & 28,600 & 3.5 \\
Put Bottle & 14,665 & 200 & 45,800 & 4.5 \\
Insert Tube & 9,427 & 200 & 29,400 & 3.6 \\
\bottomrule
\end{tabular}
\end{table}

In the second stage, we freeze the exponential moving average weights of the generator and train a SimpleTouch action policy for 20,000 updates.
Generated tactile images use the RGB channel order and tensor layout of measured images.
We use task-specific statistics to map them to the T3 input range, resize the shorter side to 256 pixels, and take a $224\times224$ center crop.
The action policy takes generated current touch as input. Latent features of future measured touch are used during training, with auxiliary loss weight 0.01.

For real-world TacImag, the generator and action policy share the current wrist image and the state input.
The generator predicts tactile images for both fingers at the observation time.
The action policy also receives the generated touch and the instruction. Its targets are the 50 demonstrated commands starting from $a_t$.
The real-world generators retain the output resolution, fourfold downsampling, U-Net channel widths, and diffusion step counts described above, and train for 200 epochs.
The generators for Ball Sorting, Insert USB, Wipe Board, and Chip Moving take 75,000, 61,000, 72,600, and 86,800 updates, respectively.
The second stage trains for 20,000 updates with future-touch auxiliary loss weight 0.01.

\subsubsection{Mixed-Action Training for HapticVLA}

Our HapticVLA baseline follows the student-stage action mixing in the official Crab implementation~\cite{hapticvlacode}.
A frozen, task-trained SimpleTouch teacher precomputes a 50-action chunk for each demonstration observation using ten flow-matching steps.
For each action step in the chunk, we average the demonstration target and teacher prediction elementwise in the eight action dimensions, $A_{\mathrm{mix}}=0.5A_{\mathrm{GT}}+0.5A_{\mathrm{teacher}}$, then zero-pad to 32 dimensions.

The student inherits the teacher's shared vision-language backbone and action weights, with the tactile modules removed.
We use the precomputed mixed actions as supervision, restart the optimizer and scheduler, and train for 20,000 updates with a global batch of 64.
The learning rate warms up to \(5\times10^{-5}\) over 1,000 updates and then decays along a cosine schedule to \(5\times10^{-6}\).

\subsubsection{Published Benchmark Results}

The ACT, VITaL, UniVTAC-ACT, and FTP-1 success rates in Table~\ref{tab:main}(a) come from the SimpleTouch benchmark study~\cite{simpletouch} on the same four simulation tasks.

\subsection{Experimental Platforms, Demonstrations, and Evaluation}
\label{sec:evaluation_protocol}

\subsubsection{Simulation Platform and Tasks}
\label{sec:simulation_details}

The four main UniVTAC tasks are Lift Bottle, Lift Can, Put Bottle, and Insert Tube.
We use the UniVTAC environment snapshot dated 2026-09-13 with Isaac Sim 4.5.0 and TacEx 0.1.0.
Cameras use high-quality rendering with FXAA. The head and wrist cameras provide native RGB images at $640\times360$, and the native tactile resolution is $400\times300$.
Lift Bottle and Put Bottle use the head view. Lift Can and Insert Tube also use the wrist view.

All methods we evaluate in UniVTAC use its default joint-position interface.
The interface sets the simulated joint state directly to the target without dynamic tracking, then advances one physics step of $1/120$\,s.
All $\pi_{0.5}$-based methods in our simulation evaluation predict from the current observation using ten flow-matching steps, without observation history or temporal ensembling.
The episode budgets count action calls, as shown in Table~\ref{tab:supp_episode_limits}.

\begin{table}[t]
\centering
\caption{\textbf{Episode action-call limits for the main simulation tasks.}}
\label{tab:supp_episode_limits}
\begin{tabular}{lr}
\toprule
Task & Action-call limit \\
\midrule
Lift Bottle & 500 \\
Lift Can & 300 \\
Put Bottle & 300 \\
Insert Tube & 300 \\
\bottomrule
\end{tabular}
\end{table}

Each simulation condition schedules 100 evaluation episodes per method and task.

\begin{figure*}[t]
\centering
\includegraphics[width=\textwidth]{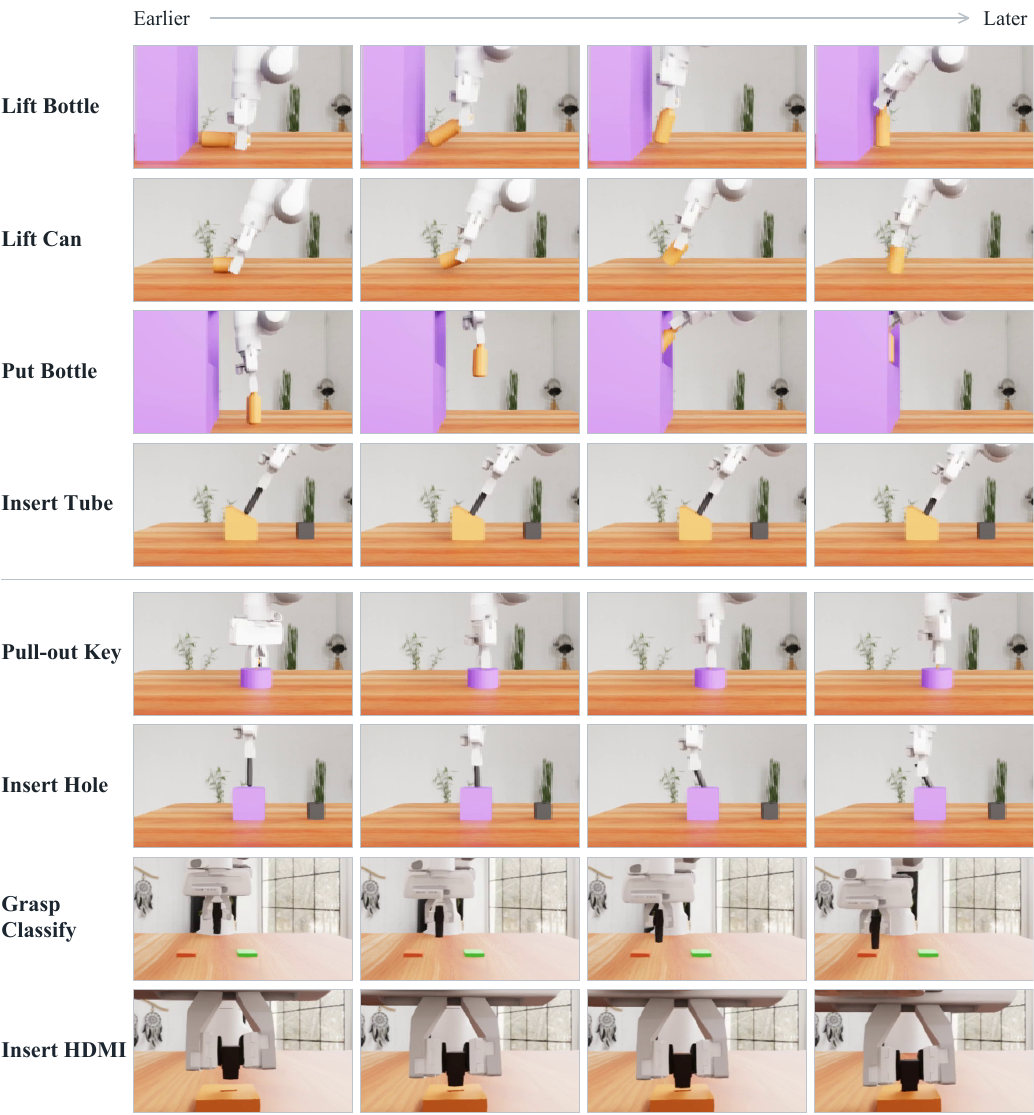}
\caption{\textbf{TacitTouch rollouts on eight UniVTAC tasks.} Each row shows four frames from one successful evaluation episode. The top four rows show the main simulation tasks, and the bottom four show the additional tasks. The fixed-camera views are cropped around the workspace.}
\label{fig:supp_univtac_rollouts}
\end{figure*}

We use the task-specific UniVTAC success criteria.
For Insert Tube, we verify lateral alignment in the recorded successful episodes using the absolute offsets $|x|<5$\,mm and $|y|<5$\,mm in the hole frame.
The task also checks insertion depth, orientation, and relative grasp height.
Section~\ref{sec:scope_details} gives the criteria for the additional tasks.
Figure~\ref{fig:supp_univtac_rollouts} shows execution sequences from the main and additional tasks.

\subsubsection{Real-World Platform and Tasks}

\begin{figure*}[t]
\centering
\includegraphics[width=\textwidth]{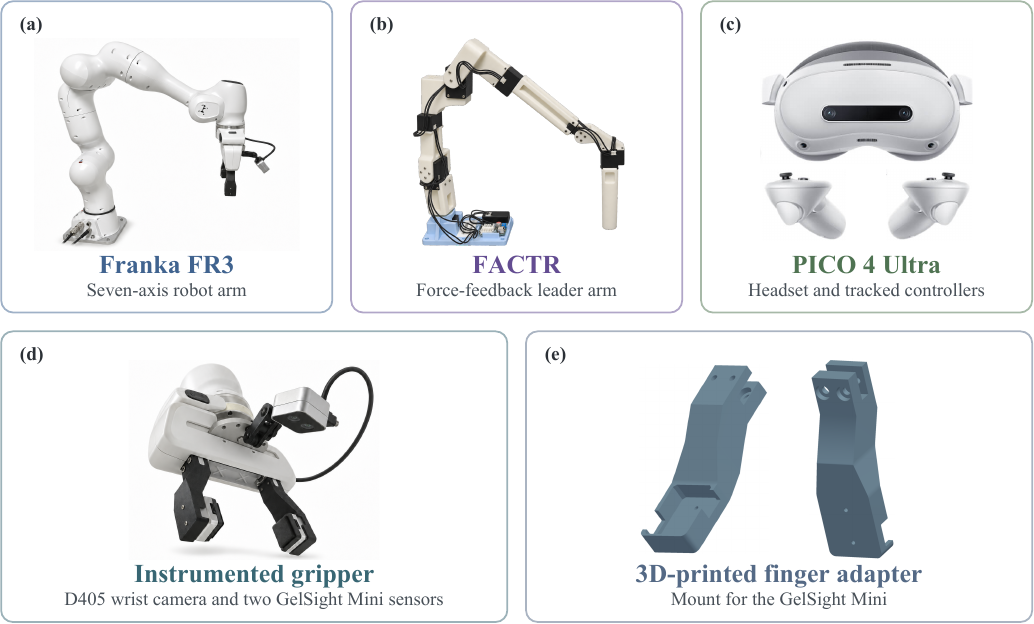}
\caption{\textbf{Hardware for real-world experiments.} (a) Franka FR3 arm. (b) FACTR leader arm for force-feedback teleoperation. (c) PICO 4 Ultra headset and tracked controllers. (d) Instrumented gripper with a RealSense D405 wrist camera and two GelSight Mini sensors. (e) 3D-printed finger adapter for the GelSight Mini.}
\label{fig:supp_franka_setup}
\end{figure*}

The Franka FR3 uses an Intel RealSense D405 on the wrist and a GelSight Mini on each gripper finger.
Two external RealSense D435i cameras record side and overhead views.
The deployed policies use the wrist image as visual input, with tactile inputs listed in Table~\ref{tab:supp_inputs}.
Figure~\ref{fig:supp_franka_setup} shows the arm, sensing hardware, and teleoperation devices.
Table~\ref{tab:supp_real_tasks} defines the four tasks.

\begin{table}[t]
\centering
\caption{\textbf{Real-world task objectives.}}
\label{tab:supp_real_tasks}
\begin{tabular}{@{}p{.26\columnwidth}p{.68\columnwidth}@{}}
\toprule
Task & Objective \\
\midrule
Ball Sorting & Pick up the ball and sort it into the corresponding container by surface texture (smooth or rough). \\
Insert USB & Grasp the USB plug, align it, and insert it into the target port. \\
Wipe Board & Pick up the eraser, make contact with the board, and erase the designated writing. \\
Chip Moving & Move the chip from the source bottle mouth to the target bottle mouth and release it so it rests intact. \\
\bottomrule
\end{tabular}
\end{table}

\subsubsection{Demonstration Collection and Processing}
\label{sec:real_demonstration_processing}

Each real-world task uses 50 teleoperated demonstrations, indexed 0--49 after conversion.
Ball Sorting uses 25 smooth-ball and 25 rough-ball demonstrations.
During teleoperation, we record three camera streams, tactile images from both fingers, measured arm and gripper state, and joint and gripper commands.

For export, we construct a regular 15\,Hz grid with an interval of approximately 66.7\,ms.
At each grid time, we take the latest available frames from the three cameras (wrist, side, and overhead) and two tactile sensors, and the latest measured arm and gripper state.
For each of the two command streams, we take the earliest command at or after the grid time.
The \texttt{state} thus contains the most recent measured feedback, and every \texttt{action} traces back to its source \texttt{command}.
Camera source timestamps use the host's monotonic clock at the end of each frame read.

\begin{figure*}[t]
\centering
\includegraphics[width=\textwidth]{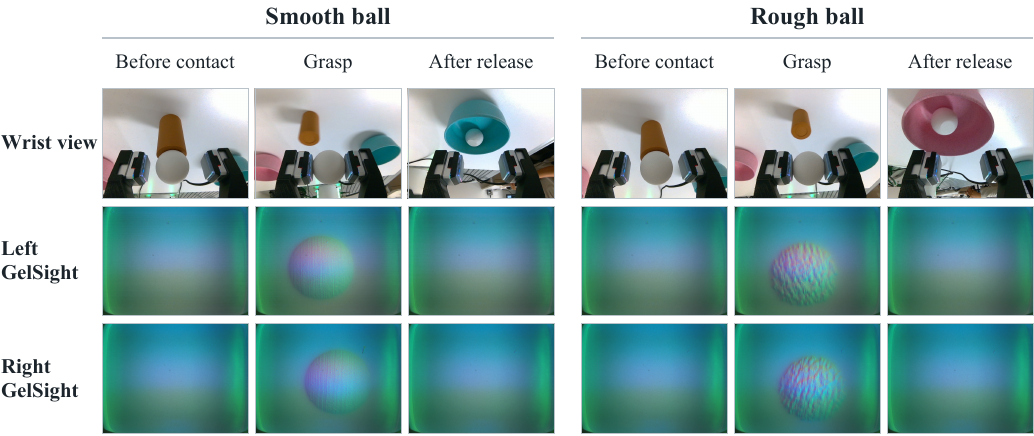}
\caption{\textbf{Contact during real-world Ball Sorting.} Aligned wrist and left/right GelSight images from a smooth-ball and a rough-ball demonstration at three stages. Each column uses the wrist image and the two measured tactile images from the same aligned sample.}
\label{fig:supp_ball_sorting_touch}
\end{figure*}

Each image stream contributes fresh frames only, with a maximum frame age of one grid interval.
We also limit the wait for each command to one interval.
We allow the latest state feedback to be up to 1\,s old.
The maximum observed gripper-feedback age is 44.35\,ms for Ball Sorting, 100.41\,ms for Insert USB, 46.14\,ms for Wipe Board, and 129.76\,ms for Chip Moving.
Alignment selects source records directly, without interpolation. Rows that fail the matching criteria stay on the grid with an invalid mask.
Figure~\ref{fig:supp_ball_sorting_touch} shows aligned wrist and tactile frames through grasp and release in Ball Sorting.

Wrist images retain their native $640\times480$ field of view.
We resize the full tactile image to $320\times240$ by area resampling, keep the RGB channel order, and encode it at JPEG quality 95.
For each task, we estimate tactile statistics using both fingers' images from all 50 training demonstrations.
For T3, we resize the shorter side to 256 pixels and take a $224\times224$ center crop.

\subsubsection{Real-World Action Execution}

All four tasks send the seven absolute joint-position targets obtained in Sec.~\ref{sec:supp_state} as references to their low-level arm controllers.
Ball Sorting passes these targets to the Franka SDK's joint-position interface.
Wipe Board uses the follower-arm controller from FACTR, a bilateral force-feedback teleoperation system.
It tracks the targets with a joint-space proportional--derivative torque law based on scaled position error and measured joint velocity.

Insert USB and Chip Moving reuse the controller from our Pico teleoperation setup, which records demonstrations using a PICO headset and a tracked handheld controller.
The controller computes the target end-effector pose by forward kinematics and uses the SDK's desired joint state to compute pose and posture errors.
It maps proportional pose feedback to joint velocities through a damped Jacobian inverse and adds a projected joint-posture correction. Ruckig smooths the resulting joint-velocity reference before execution.
The gripper uses the Move interface to execute the total opening specified by the eighth action dimension.

\subsubsection{Success Rates}

Under the simulation protocol in Sec.~\ref{sec:setup}, the success rate for a task is $\hat p=S/n$, where $S$ is the number of successful episodes and $n$ is the number of episodes included in the reported score.

For cross-task simulation means in the main paper, we round each task's success rate to an integer percentage before averaging the four rounded rates equally.
On the robot, each method runs 20 trials per task under Clean and 20 under Light.
Each set of 20 Ball Sorting trials includes ten smooth-ball and ten rough-ball trials.
Tables~\ref{tab:main}(b) and~\ref{tab:real_robot_light} report the results.

\subsection{Visual Perturbations and Matched Examples}
\label{sec:perturbation_reproducibility}

\subsubsection{Visual Perturbation Protocol}

Section~\ref{sec:robustness} evaluates the Light and Camera conditions at fixed checkpoints.
Light multiplies the RGB values of the policy camera images by 0.5 before preprocessing. The tactile brightness multiplier stays at 1.0.
We use the same input transform in simulation and on the robot. Scene illumination and physical contact conditions remain fixed.

Camera is a simulation-only condition.
We sample the camera perturbations before evaluation.
The head-camera yaw is uniform over $[-30^\circ,30^\circ]$ about the world $Z$ axis through the workspace point $(0.70,0,0.04)$\,m.
Its position follows a horizontal arc and its orientation rotates by the same angle. We add no independent random translation.
For tasks with a wrist view, we independently sample yaw uniformly over $[-15^\circ,15^\circ]$ about that camera's local $Y$ axis, with the mount position fixed.
We apply each perturbation after the environment reset and use the same perturbed views for corresponding episodes across methods.
Evaluation retains views that occlude the target, such as those in Put Bottle where the shelf opening is hidden.

\subsubsection{Selection and Display of Matched Episodes}

Figure~\ref{fig:behavior_perturbations} pairs TacImag and full TacitTouch trajectories with the same initial scene and visual perturbation.
We select one pair per condition in which TacImag fails and TacitTouch succeeds, before inspecting the frames.
The Insert Tube example uses head- and wrist-camera yaw offsets of $+26.7^\circ$ and $-12.5^\circ$, respectively.

We show the first and last decoded frames and one intermediate frame from each recording.
The reference images above the sequences use the head camera, with decoded RGB values multiplied by 0.5 in the dimmed example.
Trajectory frames retain their recorded brightness because Light applies only to policy inputs.
Lift Bottle shows the head view, and Insert Tube shows the wrist view.

The video writer resizes the native $640\times360$ image to $320\times320$. We restore the $16{:}9$ aspect ratio for display.
Both methods use the same crop and magnification within each task.

\subsection{Inference Latency, Parameters, and Memory}
\label{sec:timing}

For Table~\ref{tab:latency}, we measure the latency of a single policy call on an idle NVIDIA A100-SXM4-80GB GPU with PyTorch 2.7.1, CUDA 12.6, bfloat16, and 16 CPU threads.
Inputs come from ten decoded demonstration observations and use batch size one.
Timing starts at input conversion, includes vision-language processing, each method's tactile modules, and ten action denoising steps, and ends when the predicted 50-action chunk returns to the CPU.
We synchronize the GPU before and after each call.
Checkpoint loading, file decoding, simulation, and RPC communication fall outside the timed interval.

Warm-up proceeds in groups of ten calls, with a minimum of 20 and a maximum of 100 calls.
It stops when the relative difference between the median latencies of two consecutive groups is at most $5\%$.
In practice, warm-up stopped after 20--30 calls for every method--task setting.
We measure five rounds of 20 calls, for 100 measured calls per method and task.
Table~\ref{tab:supp_latency} reports average latency across tasks, deployment parameter counts, and the peak allocated memory measured by \texttt{torch.cuda.max\_memory\_allocated}.

\begin{table}[t]
\centering
\caption{\textbf{Inference latency and deployment cost.} Vision + Tac. denotes measured-touch SimpleTouch, as in Table~\ref{tab:latency}.}
\label{tab:supp_latency}
\begin{tabular}{lrrr}
\toprule
Method & \shortstack{Latency\\(s/call)} & \shortstack{Parameters\\(billions)} & \shortstack{Peak memory\\(GiB)} \\
\midrule
Vision + Tac. & 0.37209 & 3.98 & 7.62 \\
TacImag & 0.45363 & 4.25--4.28 & 8.13--8.18 \\
TacitTouch & 0.33660 & 3.35 & 6.45 \\
\bottomrule
\end{tabular}
\end{table}

The $\pi_{0.5}$ backbone in SimpleTouch and TacitTouch uses three fixed image slots.
We fill absent views with zero images and mask their positions in attention. The total number of image tokens and the parameter count are constant across view configurations.
At two decimal places, the per-task peak memory is 7.62\,GiB for SimpleTouch and 6.45\,GiB for TacitTouch on all four tasks.
TacImag's generator uses a dedicated visual encoder for each available view.
Lift Bottle and Put Bottle use the head image and give the lower bounds in Table~\ref{tab:supp_latency}.
Lift Can and Insert Tube use both head and wrist images and give the upper bounds.

Using unrounded mean latencies, TacitTouch reduces latency by $25.8\%$ relative to TacImag and by $9.5\%$ relative to Vision + Tac.
The simulation timing benchmark uses a GT0 student with feature distillation weight 0.01.
The main performance experiments use a GT1 student with the same deployment architecture.

\subsection{Ablation Settings}
\label{sec:ablation_details}

The component ablations in Fig.~\ref{fig:ablations} share a frozen teacher and train for 20,000 distillation updates.
Removing feature distillation retains $\mathcal{L}_{\mathrm{AD}}$ and $\mathcal{L}_{\mathrm{GT}}$.
Removing demonstration supervision retains $\mathcal{L}_{\mathrm{AD}}$ and $\mathcal{L}_{\mathrm{FD}}$.
Table~\ref{tab:plotted_values} reports per-task success rates.

Both initialization ablations in Fig.~\ref{fig:ablations} use the full objective $\mathcal{L}_{\mathrm{AD}}+0.01\mathcal{L}_{\mathrm{FD}}+\mathcal{L}_{\mathrm{GT}}$.
The visual initialization starts from task-trained $\pi_{0.5}$ weights.
The visuotactile initialization inherits the weights shared with the task's SimpleTouch teacher and removes its tactile modules.
Both restart the optimizer and learning-rate scheduler. The 20,000-update budget counts only the distillation stage.

In Table~\ref{tab:modality}, V denotes images, language, and proprioception, and VT adds measured touch.
The three GT0 students initialize their vision-language backbone and action expert from pretrained $\pi_{0.5}$ and share the objective $\mathcal{L}_{\mathrm{AD}}+0.01\mathcal{L}_{\mathrm{FD}}$.
V$\to$V distills from a visual $\pi_{0.5}$ teacher trained on the task.
VT$\to$V uses the task-trained SimpleTouch teacher and keeps the visual student architecture fixed.
VT$\to$VT uses that visuotactile teacher and adds a randomly initialized tactile expert to the student, which receives measured touch during both training and inference.
The last VT$\to$V row uses the full GT1 objective from Table~\ref{tab:supp_training}.

\subsection{Additional Simulation Tasks and Success Criteria}
\label{sec:scope_details}
\label{sec:scope}

Table~\ref{tab:scope} and Sec.~\ref{sec:transfer_scope} report transfer to Pull-out Key, Insert Hole, Grasp Classify, and Insert HDMI.
We train separate policies on 50 demonstrations per task and evaluate the checkpoint at update 20,000.

All four tasks use the head camera and the eight-dimensional state in Sec.~\ref{sec:supp_state}.
We use the simulation snapshot in Sec.~\ref{sec:simulation_details}.
Inference uses ten flow-matching steps with the current observation and no temporal ensembling.
Each additional task contributes 100 evaluation episodes to its reported success rate.

The environment applies the following task-specific success criteria.

\begin{itemize}
    \item \textbf{Pull-out Key.} The key height exceeds 0.09\,m. The environment also checks the orientation of the key relative to the slot and the end-effector height relative to the key.
    \item \textbf{Insert Hole.} The absolute relative $x$ and $y$ coordinates are both below 0.01\,m and the relative $z$ coordinate is below $-0.04$\,m. The environment also checks orientation and the change in relative grasp height.
    \item \textbf{Grasp Classify.} The absolute relative position components are below 0.02\,m, 0.02\,m, and 0.01\,m, respectively. The environment also checks orientation.
    \item \textbf{Insert HDMI.} The environment checks relative height, end-effector height, and orientation. The lateral position check uses the $y$ component alone.
\end{itemize}

\subsection{Per-Task Success Rates and Complete Ablation Results}
\label{sec:plot_values}

Table~\ref{tab:supp_task_rates} reports the integer-rounded success rates for full TacitTouch on the four main simulation tasks under Clean, Light, and Camera.

\begin{table}[!ht]
\centering
\caption{\textbf{Per-task success rates for full TacitTouch (\%).}}
\label{tab:supp_task_rates}
\begin{tabular}{llr}
\toprule
Condition & Task & Rate (\%) \\
\midrule
Clean & Lift Bottle & 96 \\
& Lift Can & 85 \\
& Put Bottle & 37 \\
& Insert Tube & 92 \\
\midrule
Light & Lift Bottle & 95 \\
& Lift Can & 86 \\
& Put Bottle & 41 \\
& Insert Tube & 88 \\
\midrule
Camera & Lift Bottle & 92 \\
& Lift Can & 64 \\
& Put Bottle & 20 \\
& Insert Tube & 58 \\
\bottomrule
\end{tabular}
\end{table}

Table~\ref{tab:robustness_delta} first rounds each task-condition success rate to an integer percentage.
It then subtracts the rounded Clean rate for each task and takes the equal-weight mean of the four differences, rounded to one decimal place.
For TacitTouch, the rounded Clean rates are $96\%$, $85\%$, $37\%$, and $92\%$. Under Camera they are $92\%$, $64\%$, $20\%$, and $58\%$.
The differences are $-4$, $-21$, $-17$, and $-34$ percentage points, for a mean of $-19.0$ percentage points.

\begin{table}[!ht]
\centering
\caption{\textbf{Success rates for each setting in Fig.~\ref{fig:ablations} (\%).} Means weight the rounded task rates equally.}
\label{tab:plotted_values}
\setlength{\tabcolsep}{2.5pt}
\begin{tabular}{@{}p{.40\columnwidth}rrrrr@{}}
\toprule
Configuration & \shortstack{Lift\\Bottle} & \shortstack{Lift\\Can} & \shortstack{Put\\Bottle} & \shortstack{Insert\\Tube} & Mean \\
\midrule
\raggedright Without feature distillation & 93 & 75 & 37 & 55 & 65.0 \\
\raggedright Without demonstration supervision & 95 & 72 & 38 & 93 & 74.5 \\
\raggedright Task-trained visual initialization & 96 & 60 & 37 & 92 & 71.3 \\
\raggedright Task-trained visuotactile initialization & 97 & 76 & 37 & 95 & 76.3 \\
\raggedright Full TacitTouch & 96 & 85 & 37 & 92 & 77.5 \\
\bottomrule
\end{tabular}
\end{table}

\subsection{Real-World Demonstrations and Visual Attention}
\label{sec:attribution}

\subsubsection{Task Stages and Two-Finger Touch}

Figure~\ref{fig:real_robot_tasks} shows teleoperated demonstrations for four tasks in execution order.
Each wrist image and the two adjacent measured tactile images come from the same aligned sample, with L and R denoting the left and right gripper fingers.
Images retain their original RGB channel order and colors.
The purple insets crop and enlarge the local interaction regions from the original images.

\subsubsection{Attention Extraction and Display}

Figure~\ref{fig:visual_sensitivity} shows attention from action queries to regions of the wrist image.
The models predict 50 actions with ten denoising steps.
At the final step, we take the attention probabilities after masking and softmax. We average over all attention heads, the first two future action queries, and the last four action-expert layers to obtain a $16\times16$ patch map.

The $\pi_{0.5}$, TacImag, and TacitTouch policies use the same wrist frame, instruction, and initial action noise.
For this figure, each policy retains its native state input.
TacImag encodes the frozen generator's tactile RGB output with T3.
All three run without measured touch.

We resize the $640\times480$ wrist image while retaining its aspect ratio, then pad the top and bottom by 28 pixels to form the $224\times224$ encoder input.
For display, we remove patch rows fully within the padding.
We express attention density relative to a uniform distribution on the visible image and bilinearly interpolate it to the original frame.
All tasks and models share a color scale capped at 15.3 times the uniform density. Values above this cap are clipped to the top color.

The Wipe Board inset enlarges writing still visible in the original wrist image.

\end{document}